\documentclass[11pt]{article}

\usepackage[preprint]{acl}

\usepackage{times}
\usepackage{latexsym}

\usepackage[T1]{fontenc}

\usepackage[utf8]{inputenc}

\usepackage{microtype}

\usepackage{inconsolata}

\usepackage{graphicx}
\usepackage{amsmath}
\usepackage{amssymb}
\usepackage[normalem]{ulem}
\usepackage{multirow}
\usepackage{makecell}
\usepackage[table]{xcolor}
\newcommand{\gray}{\cellcolor{gray!10}}
\usepackage{booktabs}
\usepackage{amsmath}
\usepackage{array}
\usepackage{arydshln}

\title{Orchestrating Intelligence: Confidence-Aware Routing for Efficient Multi-Agent Collaboration across Multi-Scale Models}

\author{Jingbo Wang$^{1}$, Sendong Zhao$^{1}$\thanks{~~Corresponding Author}, Jiatong Liu$^{1}$, Haochun Wang$^{1}$, Wanting Li$^{2}$, Bing Qin$^{1}$, Ting Liu$^{1}$\\
$^{1}$Research Center for Social Computing and Information Retrieval,\\Harbin Institute of Technology, China\\  
$^{2}$ the Institute of Automation of the Chinese Academy of Sciences, China\\
\texttt{\{jingbowang,sdzhao\}@ir.hit.edu.cn}}

\begin{document}
\maketitle
\begin{abstract}
While multi-agent systems (MAS) have demonstrated superior performance over single-agent approaches in complex reasoning tasks, they often suffer from significant computational inefficiencies. 
Existing frameworks typically deploy large language models (LLMs) uniformly across all agent roles, failing to account for the varying cognitive demands of different reasoning stages. 
We address this inefficiency by proposing OI-MAS framework, a novel multi-agent framework that implements an adaptive model-selection policy across a heterogeneous pool of multi-scale LLMs.
Specifically, OI-MAS introduces a state-dependent routing mechanism that dynamically selects agent roles and model scales throughout the reasoning process.
In addition, we introduce a confidence-aware mechanism that selects appropriate model scales conditioned on task complexity, thus reducing unnecessary reliance on large-scale models.
Experimental results show that OI-MAS consistently outperforms baseline multi-agent systems, improving accuracy by up to 12.88\% while reducing cost by up to 79.78\%.
\end{abstract}

\section{Introduction}

The rise of LLM agents in recent years has achieved great success in planning~\citep{qiao2024agent,song2023llm}, mathematical reasoning~\citep{wang2022self,swan2023math}, code generation~\citep{zhang2024codeagent,li2025codetree} and tool-augmented inference~\citep{shen2023hugginggpt,jiang2025qagent}. However, some tasks are too complex for just one ``brain.'' To solve this, researchers have started building multi-agent teams that work together, each playing a specific role like a group of human experts~\citep{hong2023metagpt,qian2024chatdev,wu2024autogen,chen2024agentverse,jiang2025cocoa}. While these teams are great at solving hard problems, they come with a major catch: they are incredibly expensive and slow. Because these agents have to talk back and forth, check each other's work, and call the LLM multiple times, the costs add up quickly. Often, these systems use a ``one-size-fits-all'' approach, using a massive, expensive LLM model for every single step, even when a smaller, faster LLM model could do the job just as well.

\begin{figure}[t]
    \centering
    \includegraphics[width=0.48\textwidth]{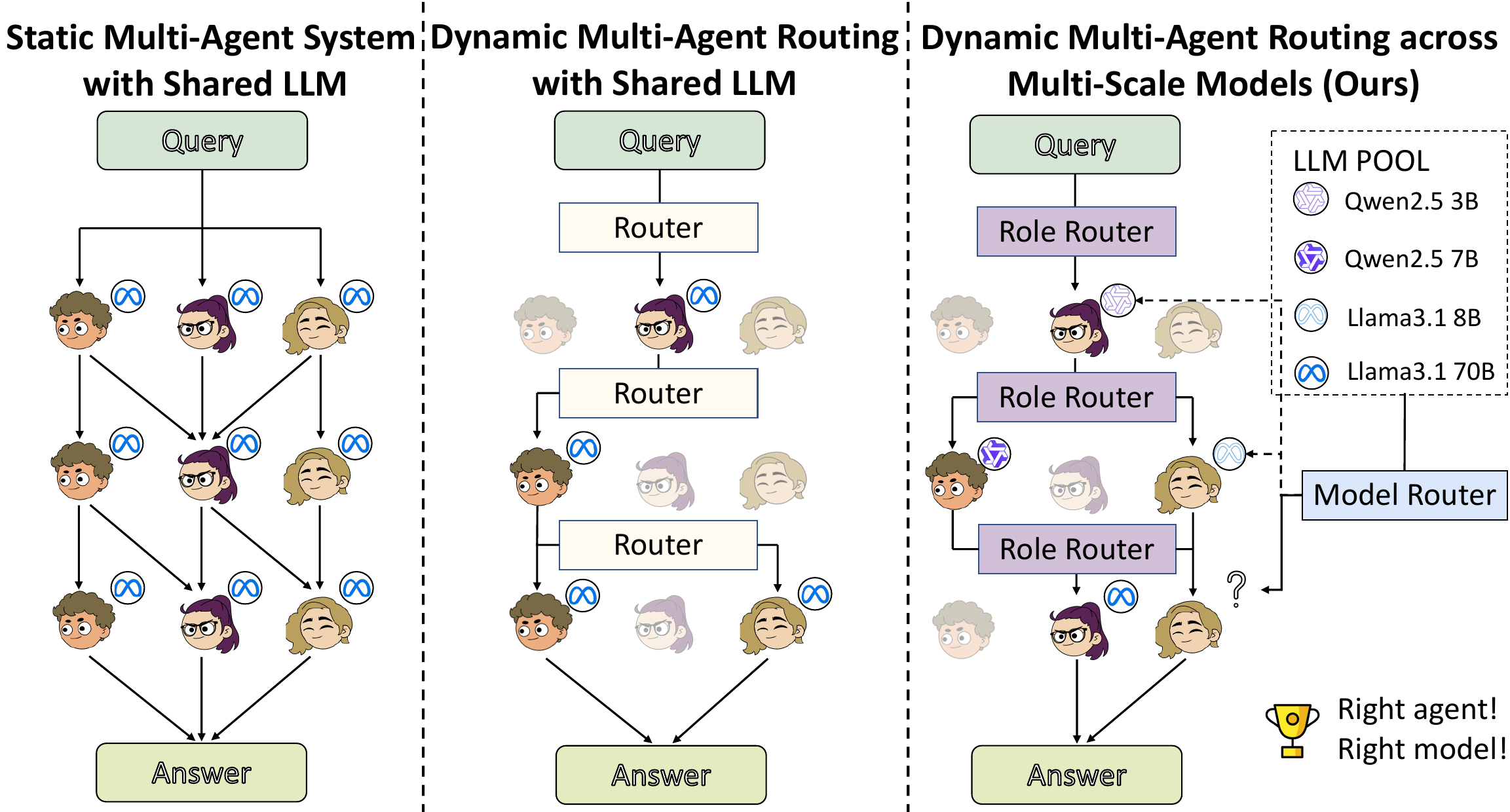}
    \caption{Paradigm comparison of static multi-agent systems, dynamic multi-agent routing with a shared LLM backbone, and dynamic multi-agent routing across a multi-scale LLM pool.}
    \label{fig:intro}
\end{figure}

So far, researchers have tried to lower these costs in two main ways. 
First, some have focused on streamlining the ``team meeting'' by optimizing how agents are organized~\citep{zhang2024g, zhuge2024gptswarm}. The second approach is ``agent routing'', where the system tries to be more selective by only calling on the specific agents needed for a certain task~\citep{dang2025multi, zhang2025multi}. These methods improve who is working and how they collaborate, however, as illustrated in Figure~\ref{fig:intro}, they overlook a fundamental issue: they tend to use the same massive, top-tier LLM for every single step. Because these systems lack flexibility in model scale, the cost savings remain limited. Even the simplest, most repetitive sub-tasks end up consuming the most expensive computational power available. In reality,  many sub-tasks in multi-agent systems are simple enough for small models (e.g., drafting or aggregation), whereas harder sub-tasks often require large-scale models (e.g., planning or revision)~\citep{ong2024routellm,ye2025x}. 
This suggests a new direction for improving multi-agent efficiency: incorporating model selection into agent routing, leading to flexible usage of different scales of LLMs.

The efficacy of this new joint model-agent routing paradigm depends on a critical challenge: \textit{How can the system know, in the middle of a complex conversation, when it needs a large-scale model and when it can get away with a smaller one}? 
Without this awareness, the system falls into two traps. It either plays it too safe and uses expensive models for everything, or it becomes overconfident and uses a small model for a critical step, causing the entire mission to fail.
To solve these problems, we introduce Orchestrating Intelligence Multi-agent System (OI-MAS), a framework that treats multi-agent reasoning like a symphony performance. In an orchestra, you don’t need every instrument playing at full volume all the time; a delicate solo might only need a flute, while a powerful climax requires the entire brass section. OI-MAS brings this same logic to AI by unifying agent roles and model scales into one dynamic process. Unlike previous systems that pick the agent and the model size separately, OI-MAS acts as a ``conductor.'' It first identifies which agent roles are needed for the current step and then selects the perfectly sized LLM ``instrument'' from a pool of different scales. To make sure the conductor makes the right call, we developed a confidence-aware mechanism. It works on a simple principle: if the system is highly confident about a task, it uses a smaller, faster model; if it senses complexity or doubt, it brings in the large-scale models. Our experiments show that this ``symphonic'' approach allows for much smarter resource use, beating existing systems in accuracy while cutting down costs and wait times significantly.

In this paper, we move away from the ``one-size-fits-all'' approach to AI teams and offer a more strategic way to build multi-agent systems. Our main contributions are:
\begin{itemize}
\item \textbf{A ``Conductor'' for AI Teams}: We propose OI-MAS, a framework that acts like a conductor in a symphony. It is the first system to jointly decide both \textbf{who} should act (the agent role) and \textbf{how much power} they need (the model scale) for every single step of a task.
\item \textbf{The Confidence-Aware Manager}: We introduce a new way to train these systems to recognize task complexity. By using \textbf{model confidence} as a signal, the system learns to ``know what it doesn't know,'' allowing it to save expensive resources for only the most challenging problems.
\item \textbf{Better Results for Less Money}: We prove through extensive testing that this approach works. By using the right model for the right job, OI-MAS doesn't just cut costs and reduce lag---it actually improves accuracy by making sure the most powerful models are focused exactly where they are needed most.
\end{itemize}

\section{Related Works}

\subsection{Multi-Agent Systems}
Multi-agent systems (MAS) have evolved from static, hand-crafted frameworks to more flexible and dynamic approaches that adapt to changing task requirements~\citep{liu2024dynamic,chen2024internet,wang2024mixture,qiu2025co,wang2025beyond}. Early methods typically employed fixed-agent teams with predefined roles and scripted interactions for each task~\citep{kim2024mdagents}. 
In contrast, more recent work has shifted toward modeling MAS as trainable graphs, where agents are treated as nodes and communication channels as edges, enabling the system to learn adaptive collaboration patterns through graph-based neural architectures~\citep{zhuge2024gptswarm, zhang2024g}.
Recent efforts have focused on automating the design of agent workflows, allowing for the generation of agentic systems without manual intervention~\citep{hu2024automated, zhang2024aflow}. 
Furthermore, orchestration and routing methods dynamically adjust agent roles and collaboration patterns based on the evolving task context~\citep{dang2025multi, zhang2025multi,wang2025optimal}. 
Despite these advancements, these methods often treat model selection as a fixed decision, failing to fully leverage multi-scale models for optimized performance and cost efficiency.

\begin{figure*}[htb]
    \centering
    \includegraphics[width=0.9\textwidth]{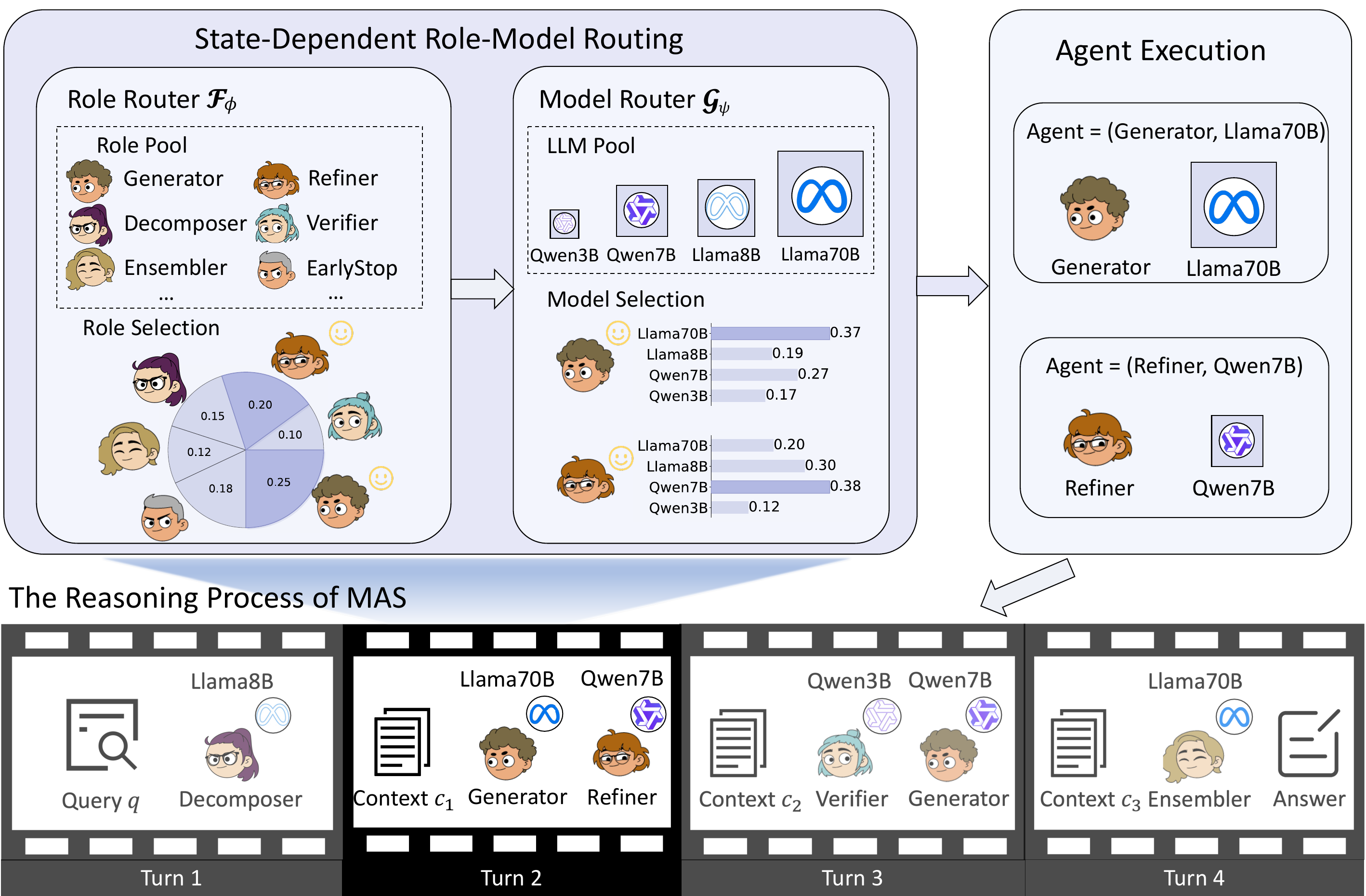}
    \caption{Overview of the proposed OI-MAS framework. The top part shows a per-turn routing policy that coordinates agent roles and assigns model capacity from a multi-scale LLM pool based on the current reasoning state; the bottom part illustrates how the system evolves across turns.}
    \label{fig:overview}
\end{figure*}

\subsection{LLM Routing}
LLM routing has been extensively studied as a principled approach to balancing predictive performance and computational cost when multiple LLMs are available. Early work addresses this via binary routing schemes, where a lightweight decision module chooses between a cheaper backbone and a more capable LLM~\citep{chen2024frugalgpt,ding2024hybrid,ong2024routellm}. 
More recent approaches extend routing to larger model pools by learning routing policies that estimate, for each input, the utility of several candidate backbones under a performance-cost objective and select the model with the highest predicted utility~\citep{lu2024routing,zhang2025capability,chen2024routerdc}, but these methods are still designed for single-agent settings. 
MasRouter brings routing into the multi-agent system by training a cascaded controller that jointly configures collaboration modes, role assignments, and LLM backbones for each task~\citep{yue2025masrouter}; however, its routing decisions are fixed before inference begins, preventing state-dependent rescheduling of agents or dynamic adjustment of models as the reasoning trajectory unfolds.
\section{Methodology}

\subsection{Preliminary}
We consider a multi-agent reasoning environment composed of a set of heterogeneous agents.
Each agent is characterized by a role that specifies its reasoning functions, along with an LLM responsible for instantiating these functions. 
We formalize the multi-agent reasoning system as
\begin{equation}\label{eq:agent-space}
\begin{gathered}
\mathcal{A} = \{\{r_i\}_{i=1}^{|\mathcal{R}|}, \{m_j\}_{j=1}^{|\mathcal{M}|}\},\\
r_i \in \mathcal{R},\quad m_j \in \mathcal{M},\\
\end{gathered}
\end{equation}
where \(r_i \in \mathcal{R}\) denotes an available role, such as generator, refiner, or programmer, and \(m_j \in \mathcal{M}\) denotes an available LLM backbone (e.g., Qwen2.5-3B, Llama3.1-70B). 

Unlike approaches that commit to a query-guided static multi-agent architecture before inference begins, our system does not assume a fixed agent pipeline. Instead, the agent configuration is evolved dynamically throughout the reasoning process. The reasoning state at turn $t$ is defined as $s_t = (q, c_t)$, where $q$ denotes the input query and $c_t$ represents the current reasoning context. Multi-agent reasoning is modeled as a discrete-time process over at most $L$ turns, where turn $t$ denotes the $t$-th interaction step during which the system selects a state-dependent subset of agents from $\mathcal{A}$, assigns each selected agent a role-model pair $(r_i, m_j)$, and executes them to transition the reasoning state from $s_t$ to $s_{t+1}$.

\subsection{State-Dependent Role-Model Routing}

To enable state-dependent routing over a pool of multi-scale models, we introduce a ``conductor'' composed of a Role Router and a Model Router, forming a hierarchical role–model routing mechanism that mirrors a two-stage process.
The first stage plans \textit{what} functional operations are required by the current reasoning state by selecting appropriate roles, abstracting away computational capacity. The second stage allocates \textit{how} these planned operations should be executed by assigning each role a backbone model whose scale matches its functional demands under a performance-cost trade-off. This decomposition decouples agent roles from resource allocation, allowing the system to plan reasoning functionality independently of model scale and to invoke larger models only when the planned operations warrant additional capacity.

\paragraph{Role Routing}
At each reasoning turn \(t\), the system determines which reasoning roles should be activated based on the current state \(s_t\).
To parameterize the routing decision, we obtain fixed semantic embeddings for the query \(q\), the current context \(c_t\), and each role description \(r_i\) through a pretrained text encoder (e.g., MiniLM~\citep{wang2020minilm}). 
These embeddings are then transformed by a learnable role network \(\mathcal{F}_\phi\), which computes a projected similarity between the representations of \((q, c_t)\) and each role \(r_i\).  After softmax normalization over all roles, the role network yields a probability distribution
\begin{equation} \label{eq:role-prob}
    p_t^{(r)}(r_i \mid q, c_t)
        = \mathcal{F}_\phi(q, c_t, r_i),
\end{equation}
The resulting probabilities are then sorted in descending order, and a subset of roles \(R_t\) is selected by accumulating probability mass until a predefined threshold $\theta$ is met, enabling the role network to activate either a single dominant role or multiple complementary roles at turn $t$, when warranted by the context.

Since some queries can be confidently resolved before all \(L\) reasoning rounds, the role space \(\mathcal{R}\) also includes a designated \textsc{EarlyStop} role that represents an explicit termination decision. When EarlyStop is included in \(R_t\), the role network concludes that additional computation is unnecessary, and the multi-agent process terminates at turn \(t\).

\paragraph{Model Routing}
Once a set of roles $R_t$ has been selected, the system assigns to each role $r \in R_t$ an appropriate model backbone from the model space $\mathcal{M}$.
Analogous to role routing, we encode the triplet $(q, c_t, r)$ into a latent representation and evaluate its suitability for every candidate backbone $m_j \in \mathcal{M}$ through a learnable model network $\mathcal{G}_\psi$. This produces a probability distribution over models:
\begin{equation}\label{eq:model-routing}
    p_t^{(m)}(m_j \mid q, c_t, r)
        = \mathcal{G}_\psi(q, c_t, r, m_j),
\end{equation}
During inference, the backbone assigned to role $r$ is obtained by selecting the model with the highest probability, yielding a role--model pairing that adapts model capacity to the evolving reasoning state.

\subsection{Confidence-Aware Optimization}

The optimization objective of our multi-agent reasoning system is to balance reasoning performance with computational cost. This balance is subtle because intermediate reasoning states differ widely in what constitutes an appropriate agent configuration. Some states are adequately addressed by a limited set of roles with smaller-scale backbones, whereas others call for larger-scale backbones, richer role compositions, or stronger inter-role interaction. Recent evidence shows that model confidence provides a reliable proxy for estimating the complexity of a reasoning state, with lower confidence indicating higher underlying complexity~\citep{zhao2025ufo}. We therefore treat confidence as a state-level indicator of agent configuration adequacy under the current routing decision.   

Building on this complexity signal, we introduce a quantitative confidence measure for each reasoning state. For a given post-decision state $\tilde{s_t} = (q, c_t, r_t, m_t)$, let $y_{1:T}$ denote the output sequence generated by the backbone model selected by the routing policy. We define the confidence of this state as the average token log-probability of the generated sequence:
\begin{equation}
\label{eq:conf}
\mathrm{Conf}_{\text{base}}(\tilde{s_t})
= \frac{1}{T} \sum_{k=1}^{T}
\log P\!\left(y_k \mid \tilde{s_t}, y_{<k}\right),
\end{equation}
This confidence score provides a signal of how well the current routing decision aligns with the requirements of the reasoning state. 
We leverage this confidence signal to modulate the cost term in a state-dependent manner, enabling the routing policy to allocate computational capacity proportionally to the estimated complexity.

\begin{table*}
\centering
\fontsize{10pt}{12pt}\selectfont
\setlength{\tabcolsep}{9pt}  
\begin{tabular}{lcccccc:c}
\Xhline{1pt}
\multicolumn{1}{l}{\textbf{Method}} & \textbf{Model} & \textbf{Gsm8k} & \textbf{MATH}  & \textbf{MedQA} & \textbf{GPQA}  & \textbf{MBPP}  & \textbf{Avg.}  \\\Xhline{1pt}
\multirow{4}{*}{Vanilla}      & Qwen2.5-3B     & 84.87          & 67.23          & 47.90           & 34.83          & 61.34          & 59.23          \\
                              & Qwen2.5-7B     & 85.71          & 72.27    & 62.18          & 33.71          & 68.07          & 64.39          \\
                              & Llama3.1-8B    & 80.67          & 53.78          & 54.62          & 32.58          & 63.87          & 57.10          \\
                              & Llama3.1-70B   & 88.24          & 67.23          & 74.79          & 35.96          & 77.31          & 68.71          \\\hline
\rowcolor{gray!10}                              
\gray
                              & Medium$^{\star}$         & 90.76          & 73.11          & 64.71          & 34.83          & 70.59          & 66.80          \\
\multirow{-2}{*}{\gray LLM-Debate}
                              &\gray Large          &\gray 94.96          &\gray 71.43          &\gray \underline{77.31}          &\gray 40.45          &\gray 82.35          &\gray 73.30    \\
\multirow{2}{*}{GPTSwarm}     & Medium$^{\star}$         & 64.71          & 52.10          & 60.50          & 37.08          & 59.66               & 54.81               \\
                              & Large          & 94.12          & 65.55          & 74.79          & 34.83          & 76.47               & 69.15               \\
\rowcolor{gray!10}
                              & Medium$^{\star}$         & 93.28          & 68.07          & 63.87          & 33.71          & 37.82          & 59.35          \\
\multirow{-2}{*}{\gray AFlow}
                              &\gray Large          &\gray 94.12          &\gray 68.91          &\gray 75.63    &\gray 42.70           &\gray 45.38          &\gray 65.35          \\
\multirow{2}{*}{MaAS}         & Medium$^{\star}$         & 88.24          & \underline{74.79}          & 55.46          & 38.20          & 79.00          & 67.14          \\
                              & Large          & \textbf{96.64} & 62.18          & 76.47          & \underline{43.82}    & \underline{88.24}    & \underline{73.47}          \\
\rowcolor{gray!10}
MasRouter                     & LLM Pool       & \underline{95.80}    & 72.27    & 71.43          & 35.96           & 77.31          & 70.55          \\\hline
OI-MAS (Ours)                          & LLM Pool       & \underline{95.80}    & \textbf{79.83} & \textbf{78.99} & \textbf{44.94} & \textbf{91.59} & \textbf{78.23} \\\Xhline{1pt}
\end{tabular}
\caption{\label{main results}
Performance comparison with vanilla models and baseline multi-agent systems. The best results are highlighted in bold, and the runner-up results are underlined.
The Medium, Large, and LLM Pool settings follow Section~\ref{subsec:Experimental Setup}, Medium$^{\star}$ marks the best result under the Medium setting.
}
\end{table*}

The confidence-aware optimization objective of the conductor is formulated as a reinforcement learning problem. Specifically, we define the objective as:
\begin{equation}
\label{eq:conf-optim-final}
\begin{aligned}
\min_{\phi,\psi}\;
&\mathbb{E}_{(q,a)\sim\mathcal{D}}\!\Big[
    -\, r(q,a;\phi,\psi)
\\
    &\quad + \sum_{t} \lambda \cdot \mathrm{Conf}_{\text{adj}}(\tilde{s_t}) \cdot C(r_t,m_t)
\Big],
\end{aligned}
\end{equation}
where $(q,a)$ denotes the input query and its ground-truth answer, and $r(q,a;\phi,\psi)\in\{0,1\}$ is a sparse reward indicating whether the final system output is correct. $C(r_t, m_t)$ measures the computational expenditure incurred by the agent choice at step~$t$. 
We further define $\mathrm{Conf}_{\text{adj}}(\tilde{s_t})\in[0,1]$ as a calibrated confidence score obtained by normalizing the raw confidence $\mathrm{Conf}_{\text{base}}(\tilde{s_t})$ to ensure semantic alignment and cross-model comparability, where larger values indicate higher confidence.
In Eq.~(\ref{eq:conf-optim-final}), $\mathrm{Conf}_{\text{adj}}(\tilde{s_t})$ acts as a state-dependent weight on the cost term, indicating whether the current role--model assignment already provides a sufficiently strong backbone model for the current state. Higher values indicate that the current routing decision is already sufficient for the state and thus enforce a stronger cost penalty to avoid unnecessary escalation to larger-scale backbone models, whereas lower values relax the cost constraint and allow rerouting to larger-scale models or richer role compositions when needed.
The computation of $\mathrm{Conf}_{\text{adj}}(\cdot)$ are detailed in Appendix~\ref{sec:appendix-confidence}.

\section{Experiments}

\subsection{Experimental Setup}
\label{subsec:Experimental Setup}
\paragraph{Datasets}
To evaluate our framework across diverse reasoning skills, we use benchmarks covering mathematics, reasoning, and programming. For mathematical reasoning, we adopt GSM8K~\citep{cobbe2021training} for multi-step arithmetic problems and MATH~\citep{hendrycks2measuring} for competition-level symbolic reasoning. For professional reasoning, we include MedQA~\citep{jin2021disease} for medical exam-style questions and GPQA~\citep{rein2024gpqa} for graduate-level physics reasoning. For programming, we evaluate on MBPP~\citep{austin2021program}, a Python function-generation benchmark assessed using pass@1.

\paragraph{Baselines}
We compare our approach with several representative multi-agent reasoning baselines. 
(1) LLM-Debate~\citep{du2023improving}: Enhances reasoning quality by enabling multiple LLMs to critique and refine one another's responses. 
(2) GPTSwarm~\citep{zhuge2024gptswarm}: Formulates LLM agents as optimizable computational graphs whose node prompts and communication edges are jointly improved. 
(3) AFlow~\citep{zhangaflow}: Automatically discovers effective agentic workflows through Monte Carlo Tree Search over code-based workflow structures. 
(4) MaAS~\citep{zhangmulti}: Optimizes a probabilistic supernet of multi-agent architectures and dynamically samples query-specific systems. 
(5) MasRouter~\citep{yue2025masrouter}: Constructs multi-agent systems by routing collaboration modes, agent roles, and LLM backbones through a cascaded controller.

\paragraph{LLM Backbones}
We adopt a heterogeneous LLM pool consisting of Qwen2.5-3B~\citep{hui2024qwen2} as the \textsc{small} model, Qwen2.5-7B~\citep{hui2024qwen2} and Llama3.1-8B~\citep{grattafiori2024llama} as \textsc{medium-scale} models, and Llama3.1-70B~\citep{grattafiori2024llama} as the \textsc{large model}. \textsc{LLM Pool} setting is defined as allowing a system to use any backbone in this pool during inference.

\begin{figure}[t]
    \centering
    \includegraphics[width=0.48\textwidth]{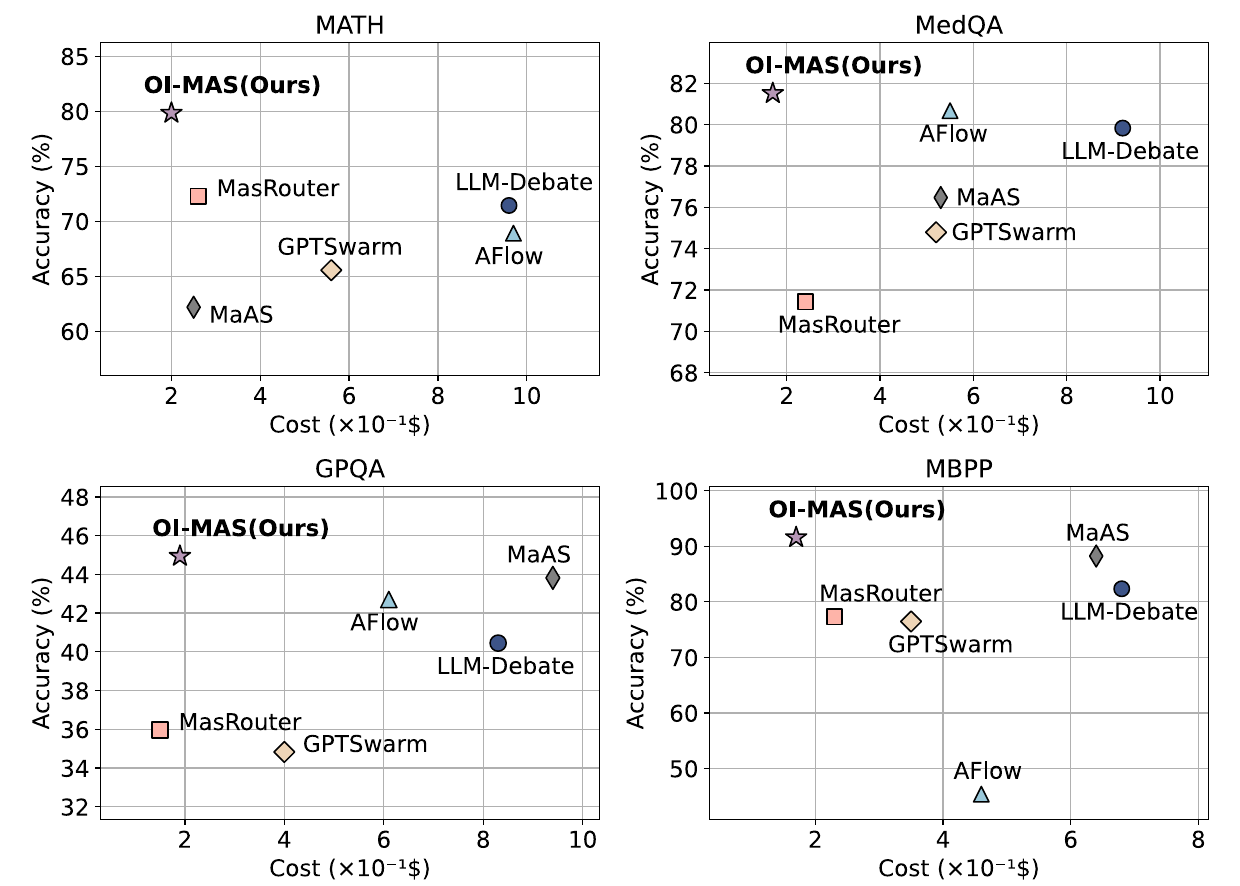}
    \caption{The comparison of accuracy and inference cost across four benchmarks, where different marker shapes denote different baseline categories.}
    \label{fig:cost}
\end{figure}
\paragraph{Implementation Details}
A diverse set of agent roles is employed, such as \textsc{Generator}, \textsc{GeneratorCoT}, \textsc{Decomposer}, \textsc{Critique}, \textsc{Ensembler}, \textsc{Verifier}, \textsc{Refiner}, \textsc{Programmer}, and \textsc{EarlyStop}, which are selectively activated on demand depending on the scenario. We set the maximum number of reasoning turns as $L = 4$ and the cost penalty coefficient as $\lambda = 200$. Role selection accumulates probability until reaching a threshold of $\theta=0.3$, and all LLM decoding is performed with temperature set to 0. The routing networks are optimized with a learning rate of $\alpha = 0.01$. All experiments are conducted on NVIDIA A100 80G GPUs with vLLM for accelerated inference, and all baselines are evaluated under identical hardware and decoding settings to ensure fair comparison.

\subsection{Main Results}

\paragraph{Superior Performance}
As reported in Table~\ref{main results}, OI-MAS consistently outperforms all vanilla backbone models of different scales across the five benchmarks, with accuracy gains ranging from 9.52\% to 21.13\%. This highlights the limitation of single-model inference, which relies on a fixed model and lacks explicit role specialization.
Beyond single-model baselines, OI-MAS also outperforms multi-agent baselines on nearly all benchmarks, demonstrating that its advantage does not stem from backbone scale alone, but from more effective coordination between agent roles and model capacities.
On MATH, strong performance is achieved under both medium and large models, and OI-MAS further delivers a clear improvement by integrating the complementary strengths of different models.
On MBPP, MaAS and OI-MAS obtain clear gains by incorporating a \textsc{Verifier} agent that exploits code executability to mitigate intermediate errors.
OI-MAS further achieves better performance, consistent with its state-dependent design that monitors intermediate states and adaptively allocates model capacity when needed.
Moreover, compared with MasRouter, which performs query-level routing over an LLM pool, OI-MAS achieves an average improvement of 7.68\% and outperforms it on four of the five benchmarks. This further underscores the benefit of state-dependent and confidence-aware routing during inference, enabling more effective utilization of multi-scale LLMs.

\begin{figure}[t]
    \centering
    \includegraphics[width=0.48\textwidth]{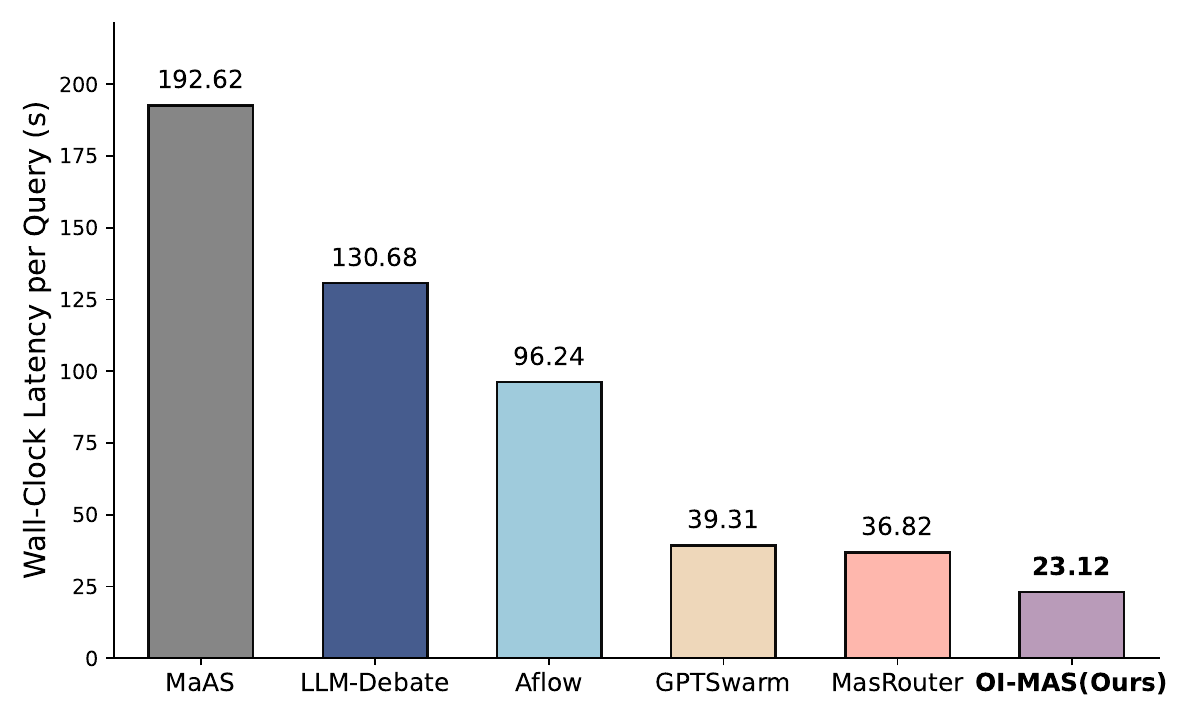}
    \caption{Wall-clock latency of OI-MAS and baselines on the GPQA benchmark.}
    \label{fig:latency}
\end{figure}

\paragraph{Cost Efficiency}
As shown in Figure~\ref{fig:cost}, OI-MAS consistently achieves a favorable accuracy-cost trade-off across all four benchmarks. Compared with all baseline multi-agent systems, OI-MAS reduces inference cost by 17.05\%--78.47\% in terms of the average inference cost over the four benchmarks, while maintaining or improving overall accuracy. 
This advantage stems from a more intelligent resource allocation strategy during inference.
Methods such as AFlow tend to rely on large-scale but expensive models to secure performance, which incurs unnecessary cost on simple tasks. While MasRouter introduces routing, its model selection is static, determined once at the beginning of a task and unable to adapt to evolving states.
In contrast, our method introduces a confidence-aware training objective, making the system inherently aware of the state's evolving complexity.
Our routing policy therefore assigns lightweight models to handle most easy subtasks and escalates to larger-scale models only when warranted, thereby avoiding redundant large-scale model invocations and substantially reducing inference cost.
The inference cost is computed using the token-based pricing scheme described in Appendix~\ref{app:cost}.

\paragraph{Reduced Latency}

OI-MAS achieves markedly lower per-query wall-clock inference latency than all compared multi-agent baselines. As shown in Figure~\ref{fig:latency}, OI-MAS completes a single inference in 23.12s, outperforming GPTSwarm (39.31s) and MasRouter (36.82s), and exhibiting an even larger advantage over more compute-intensive methods. 
This improvement is largely attributable to routing more steps to lightweight models, whose lower cost is associated with significantly shorter runtime than large-scale models.
Complementing the model selection strategy, the early-stopping behavior further reduces wall-clock latency by truncating the sequential reasoning process once a satisfactory result has been obtained, thereby avoiding additional rounds whose marginal benefits are limited.
As a result, OI-MAS shortens the end-to-end reasoning trajectory while preserving strong performance, making it more suitable for latency-sensitive multi-agent deployment scenarios.
\section{Analysis}

\begin{figure}[t]
    \centering
    \includegraphics[width=0.47\textwidth]{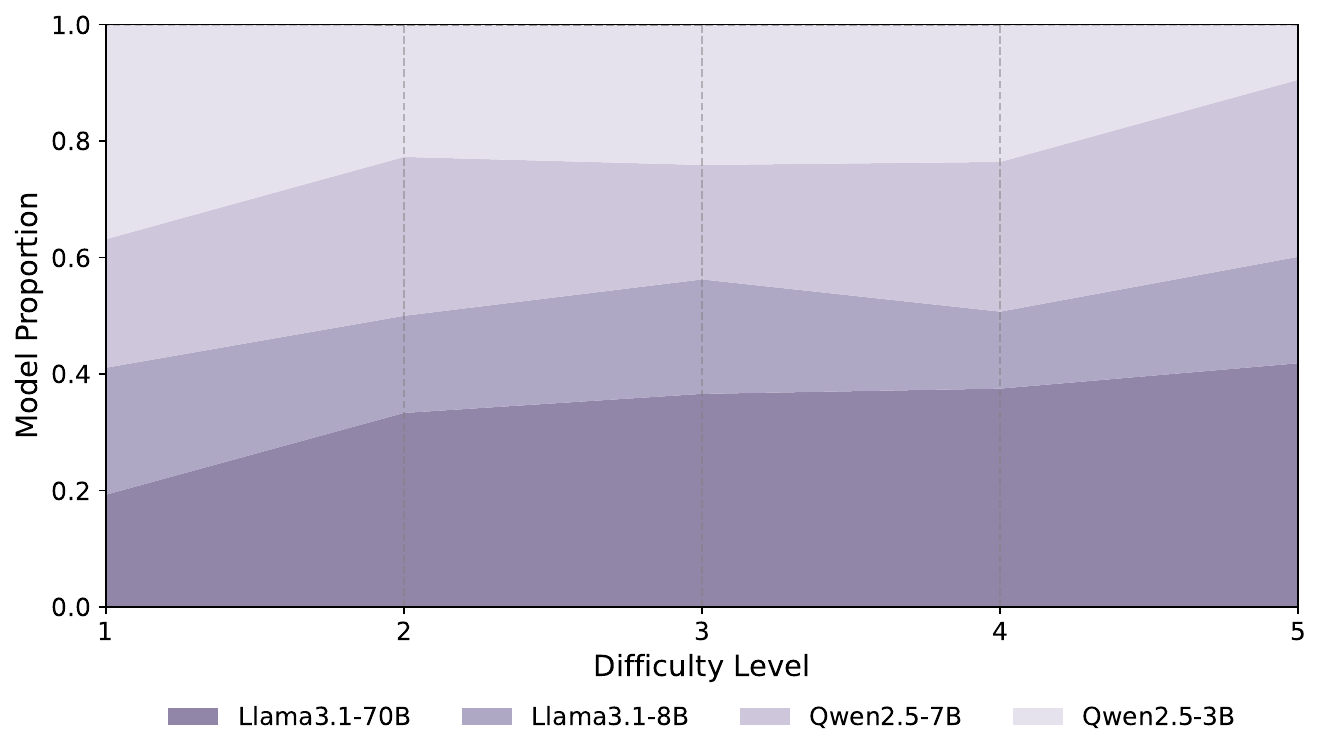}
    \caption{Model selection distribution across the five difficulty levels on the MATH dataset.}
    \label{fig:model}
\end{figure}

\subsection{Routing Behavior Analysis}
To better understand how OI-MAS adapts its routing policy to the underlying reasoning difficulty, we examine its model and role selection behaviors on the MATH dataset. The dataset provides a five-level difficulty annotation, offering a natural axis for analyzing complexity-dependent routing dynamics. In particular, we analyze how model selection varies with problem difficulty and how it differs across roles.

\paragraph{Model Selection Across Difficulty Levels}
Figure~\ref{fig:model} shows a clear progression in the distribution of selected models as MATH problem difficulty increases: OI-MAS relies less on small models and more on medium-sized and large models as the problems become harder.
This behavior suggests that the routing policy distinguishes reasoning states by their underlying difficulty. States that can be resolved with confident predictions are assigned to small models, whereas states involving greater uncertainty or more complex reasoning trajectories trigger the selection of larger models.
Such behavior indicates that the router exploits systematic regularities in the structure of multi-step reasoning: states that consistently appear in easier segments of the reasoning trajectory are routed to low-cost models, while states associated with harder segments bring higher-capacity models into play. This interaction between reasoning difficulty and model assignment shows that the routing policy adapts model scale to the demands of the evolving state, rather than relying on a fixed allocation scheme.

\begin{figure}[t]
    \centering
    \includegraphics[width=0.47\textwidth]{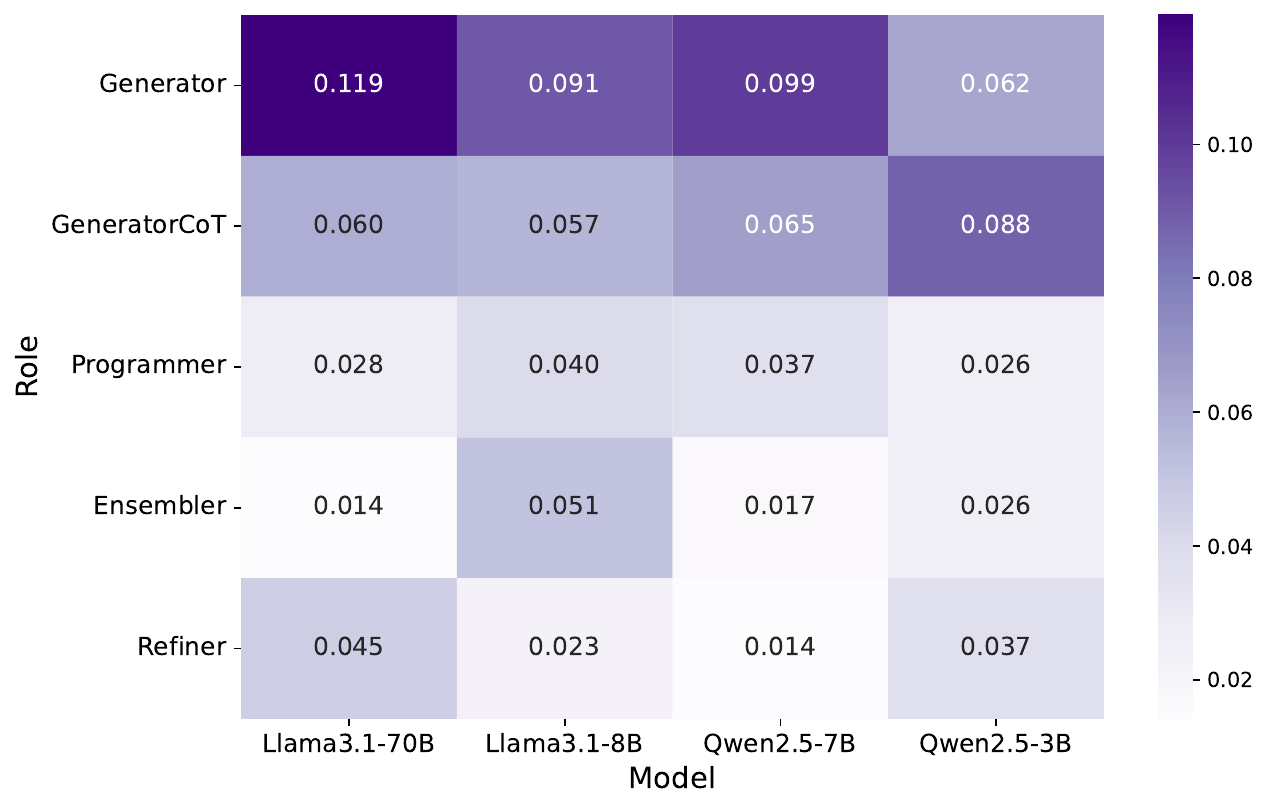}
    \caption{Model selection distribution across agent roles on the MATH dataset.}
    \label{fig:role_model}
\end{figure}

\paragraph{Role-Model Interaction Analysis}
Beyond difficulty, routing decisions also reflect the functional demands of different roles. As shown in Figure~\ref{fig:role_model}, generative roles such as \textsc{Generator} and \textsc{GeneratorCoT} exhibit a noticeably higher tendency to invoke the largest backbone compared with other roles. These roles are responsible for constructing core reasoning trajectories and producing major intermediate content, and the router tends to allocate higher-capacity models to them.
In contrast, structural and post-processing roles, including \textsc{Programmer} and \textsc{Ensembler}, are concentrated around the medium backbones, with substantially fewer escalations to the largest model. \textsc{Refiner} presents a distinct bimodal pattern: easy refinements are delegated to the smallest model, while difficult inconsistencies lead to escalation. These broader patterns indicate that OI-MAS learns role-level capacity regimes that align model selection with the functional demands of each role, thereby contributing to more efficient resource allocation within the multi-agent system.

\subsection{Out-of-Distribution Generalization}
To assess the out-of-distribution generalization ability of our approach, we evaluate OI-MAS by training the routing policy on MBPP and directly applying it to HumanEval, a benchmark for functional code generation, without any retraining. As shown in Table~\ref{tab:generalization}, OI-MAS achieves a Pass@1 of 91.46\%, surpassing the strongest baseline MaAS. 
In addition, the cost of our method is the lowest of all compared methods and less than half that of the next-best performing baseline.
This indicates that the role-model routing strategy and confidence-aware mechanism learned on MBPP transfer effectively to an out-of-distribution setting, enabling OI-MAS to maintain performance even on unseen tasks.

\begin{table}
\centering
\begin{tabular}{lccc}
\toprule
Method               & Pass@1(\%)& Cost($10^{-1}$$\$$)\\\midrule
LLM-Debate    & 78.05                                   & 2.97 \\
GPTSwarm      & 75.00                                      & 1.61 \\
Aflow         & 78.66                                   & 1.25 \\
MaAS          & \underline{89.63}                                   & 2.22 \\
MasRouter         & 74.39                                   & \underline{1.17} \\\midrule
OI-MAS (Ours)      & 91.46                                   & 0.97 \\\bottomrule
\end{tabular}
\caption{\label{tab:generalization}
Out-of-distribution evaluation on HumanEval with the routing policy trained on MBPP.
}
\end{table}

\begin{table}
\vspace{-1em}
\vspace{0.1em}
\centering
\fontsize{10pt}{13pt}\selectfont
\setlength{\tabcolsep}{3pt}  
\resizebox{\columnwidth}{!}{
\begin{tabular}{l|cc|cc}
\toprule
\makecell{Dataset} &\multicolumn{2}{c|}{MedQA} & \multicolumn{2}{c}{MBPP}\\
\midrule
\makecell{Metric}  &  \makecell{Accuracy \\(\%)} & \makecell{Cost\\ ($10^{-1}$$\$$)}   &  \makecell{Pass@1 \\(\%)} & \makecell{Cost\\ ($10^{-1}$$\$$)}  \\
\midrule
\makecell{OI-MAS} & $78.99$ & $1.79$ & $91.59$ & $1.67$ \\
\midrule
\textit{w/o} \(\mathcal{G}_\psi\)  & $81.51$ & $3.16$ & $93.28$ & $3.07$ \\s
\textit{w/o} $C(\cdot)$ & $79.83$ & $2.14$ &  $92.43$ & $1.96$   \\
\textit{w/o} $Conf(\cdot)$ & $76.47$ & $1.68$ & $87.39$ & $1.53$\\
\bottomrule
\end{tabular}}
\caption{Ablation study of OI-MAS.}\label{tab:ablation}
\vspace{-0.8em}
\end{table}

\subsection{Ablation Study}
We conduct an ablation study on three core components of OI-MAS: (1) \textbf{\textit{w/o}}~$\mathcal{G}_\psi$, which disables the model router in Equation~\ref{eq:model-routing}  and forces all agents to use the large-scale model; (2) \textbf{\textit{w/o}}~$C(\cdot)$, which removes the cost term from the optimization objective in Equation~\ref{eq:conf-optim-final}; and (3) \textbf{\textit{w/o}}~$Conf(\cdot)$, which drops the confidence-aware weighting in Equation~\ref{eq:conf-optim-final}. 
As shown in Table~\ref{tab:ablation}, removing $\mathcal{G}_\psi$ slightly improves performance but substantially increases inference cost. 
Removing $C(\cdot)$ yields a similar cost inflation with limited performance change, indicating that the explicit cost term is necessary to prevent the policy from drifting toward overly expensive configurations. 
Removing $Conf(\cdot)$ leads to a pronounced performance degradation on both MedQA and MBPP, indicating that this component plays a critical role in maintaining overall task accuracy.

\subsection{Hyperparameter Sensitivity Analysis}
\begin{figure}[t]
    \centering
    \includegraphics[width=0.47\textwidth]{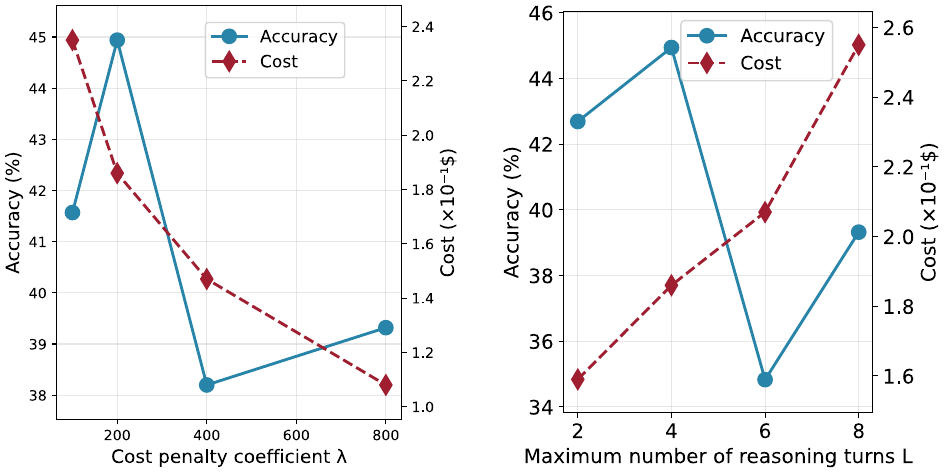}
    \caption{Sensitivity analysis of key hyperparameter on the GPQA benchmark.}
    \label{fig:hyperparameter}
\end{figure}

In Figure~\ref{fig:hyperparameter}, we analyze the sensitivity of our method to the cost penalty coefficient $\lambda$ and the collaboration turn parameter $L$ on the GPQA benchmark.
For $\lambda$, increasing it within a moderate range improves accuracy while reducing inference cost, suggesting that appropriate cost regularization suppresses unnecessary computation while simultaneously improving performance. However, overly large $\lambda$ leads to a noticeable accuracy drop, indicating that excessive regularization pushes the routing policy toward overly conservative, low-cost configurations.
For $L$, increasing collaboration turns from 2 to 4 improves accuracy with only a modest cost increase, while further increasing $L$ incurs substantial overhead and degrades performance, likely due to redundant interactions and accumulated noise.

\section{Conclusion}
In this paper, we propose OI-MAS, a dynamic multi-agent collaboration framework inspired by symphony performance. OI-MAS adopts a conductor that allocates both agent role and LLM backbone, and optimizes a confidence-aware objective to adaptively allocate agent roles and model backbones for each reasoning state. Extensive experiments demonstrate that our method selects appropriate model based on the reasoning state, achieving consistent gains in accuracy while substantially reducing computational cost. We hope this work provides a meaningful step toward more reliable and efficient multi-agent reasoning systems.


\section*{Limitations}
This study has several limitations.
First, the work does not explicitly investigate how agent memory should be represented, maintained, and governed over time, which may affect long-horizon coherence and the reliability of iterative collaboration. Second, while the experiments provide evidence of a favorable performance-cost balance under the evaluated settings, it is not guaranteed that the same balance will hold uniformly in highly concurrent, large-scale deployments. Third, agent safety is not treated as a central design objective. In multi-agent settings, interaction can introduce additional risk vectors—including unsafe tool invocation, emergent undesired behaviors, and the propagation or amplification of policy-noncompliant actions across agents—yet the paper does not present a comprehensive safety framework that is specifically calibrated to large, interactive agent collectives.

\bibliography{custom}

\appendix

\section{Dataset Description}
\label{app:datasets}

We evaluate our method on publicly available benchmark datasets spanning mathematical reasoning, professional reasoning, and programming tasks.

GSM8K is a benchmark for grade-school-level mathematical reasoning, consisting of multi-step arithmetic word problems.

MATH is a competition-level mathematical reasoning dataset covering diverse domains such as algebra, geometry, and number theory.

MedQA consists of medical examination questions designed to evaluate professional-domain reasoning.

GPQA is a graduate-level physics question answering benchmark that emphasizes deep scientific reasoning.

MBPP is a programming benchmark composed of Python function-generation tasks evaluated by functional correctness.

HumanEval is a widely used code-generation benchmark designed to evaluate the functional correctness of generated programs. Due to its relatively small size, HumanEval is used exclusively for testing and is not included in the training process.

For all datasets except HumanEval, we adopt a train/test split with a ratio of $4{:}1$.

\section{Details of Confidence Adjustment}
\label{sec:appendix-confidence}

As defined in Equation~\ref{eq:conf}, we use the average token log-probability as the raw confidence signal, denoted by $\mathrm{Conf}_{\text{base}}(\tilde{s_t})$. While $\mathrm{Conf}_{\text{base}}(\tilde{s_t})$ correlates with the local complexity of the reasoning state, it is not directly suitable for confidence-aware optimization.

The raw confidence signal suffers from two limitations. First, it is \textit{semantically inverted}: since $\mathrm{Conf}_{\text{base}}(\tilde{s_t})\le 0$, higher confidence corresponds to values closer to $0$, which is misaligned with downstream cost modulation. Second, it lacks \textit{cross-model comparability}: heterogeneous backbone models can produce log-probabilities on different numerical scales.

To address these issues, we transform $\mathrm{Conf}_{\text{base}}(\tilde{s_t})$ into an adjusted confidence score $\mathrm{Conf}_{\text{adj}}(\tilde{s_t})\in[0,1]$, where larger values consistently indicate higher confidence. For each backbone model, we apply a model-specific normalization of $\mathrm{Conf}_{\text{base}}(\cdot)$ based on running statistics over recent states (e.g., percentile-based scaling). In cold-start or low-data regimes, where such statistics are unreliable, we fall back to a bounded, model-agnostic transformation using the geometric mean token probability, $\exp(\mathrm{Conf}_{\text{base}}(\tilde{s_t}))$. The final $\mathrm{Conf}_{\text{adj}}(\tilde{s_t})$ is obtained by smoothly interpolating between the fallback and the model-specific normalization as more observations accumulate.

The adjusted confidence $\mathrm{Conf}_{\text{adj}}(\tilde{s_t})$ is incorporated into the objective in Eq.~\ref{eq:conf-optim-final} as a monotonic weight on the cost term. The weighting function is increasing in confidence, such that the effective cost penalty grows with $\mathrm{Conf}_{\text{adj}}(\tilde{s_t})$. This formulation enables confidence-aware cost discipline while resolving the semantic inversion and cross-model inconsistency of the raw log-probability signal.

\begin{figure*}[htb]
    \centering
    \includegraphics[width=0.88\textwidth]{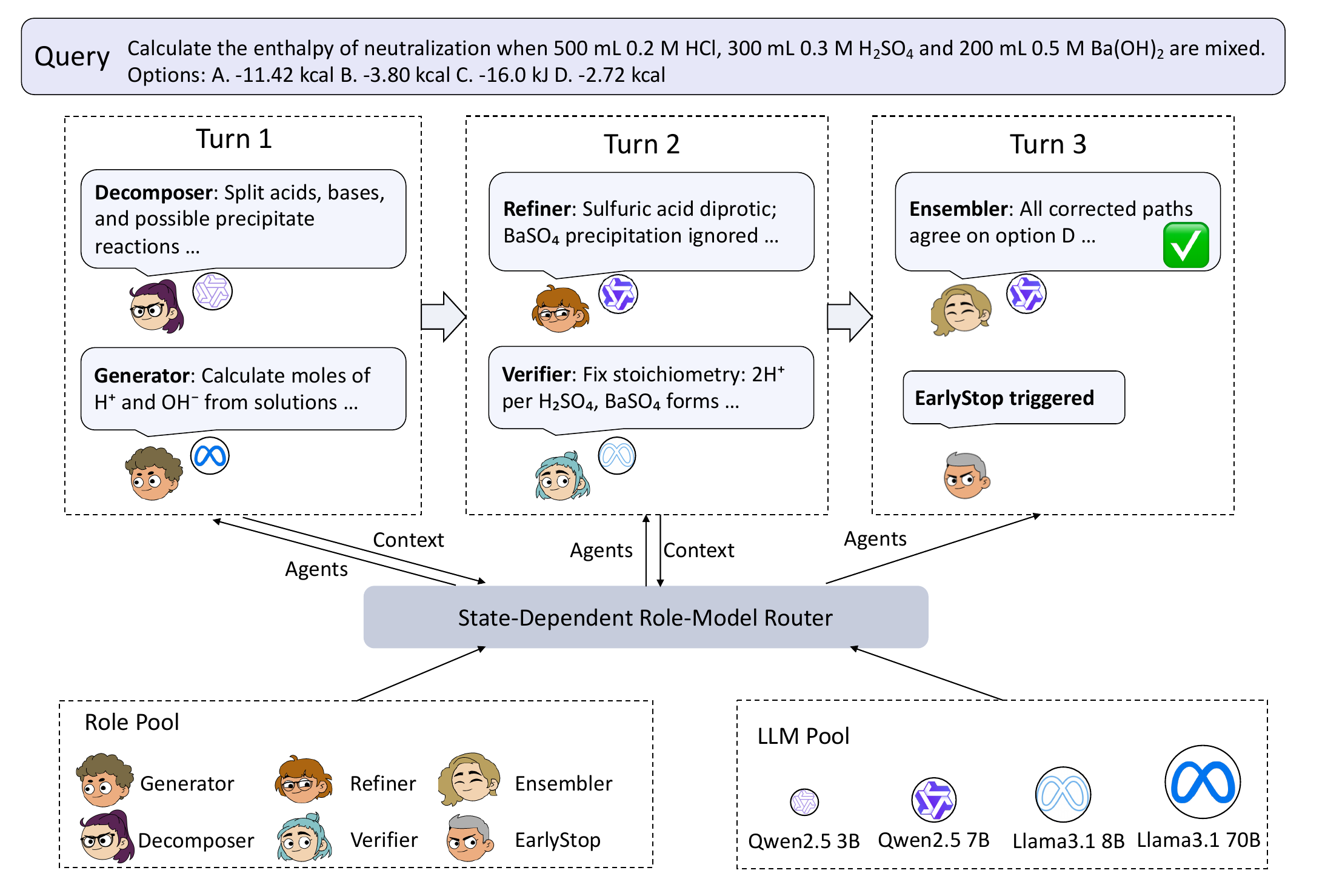}
    \caption{Case study on GPQA benchmark.}
    \label{fig:case_study}
\end{figure*}

\section{Cost Computation across Models}
\label{app:cost}

To ensure a transparent and consistent performance-cost comparison across backbone models, we adopt a unified unit token pricing scheme in all cost-related analyses. Although all backbones are deployed and executed locally in our experiments, we use API token pricing as a common proxy for inference cost to enable fair and comparable cost accounting across models.

For Llama3.1-70B, Llama3.1-8B, and Qwen2.5-7B, the unit token prices are taken from the official Together AI API pricing page.\footnote{\texttt{https://www.together.ai/pricing}} For Qwen2.5-3B, since an explicit price is not available, we estimate its unit token price via a parameter-based scaling law:
\begin{equation}
C(m) \;=\; C_{\text{base}} \cdot \left(\frac{P(m)}{P_{\text{base}}}\right)^{\alpha},
\label{eq:price_fit}
\end{equation}
where $C_{\text{base}}$ and $P_{\text{base}}$ denote the unit token price and parameter count of the base model (Qwen2.5-7B), $P(m)$ is the parameter count of model $m$, and $\alpha$ is a scaling exponent. We estimate $\alpha$ from the available pricing pair (Llama3.1-70B, Llama3.1-8B), yielding $\alpha=0.73$, and apply Eq.~\eqref{eq:price_fit} to obtain the price of Qwen2.5-3B. The final unit token prices used throughout the paper are summarized in Table~\ref{tab:model_prices}.
This setup provides a reproducible and internally consistent cost basis for routing and multi-agent cost accounting, enabling fair performance-cost comparisons across heterogeneous models.

\begin{table}
\centering
\begin{tabular}{lcc}
\toprule
\textbf{Model} & \textbf{Input}(\$) & \textbf{Output}(\$) \\
\midrule
Llama3.1-70B & 0.88 & 0.88 \\
Llama3.1-8B  & 0.18 & 0.18 \\
Qwen2.5-7B   & 0.30 & 0.30 \\
Qwen2.5-3B   & 0.16 & 0.16 \\
\bottomrule
\end{tabular}
\caption{Cost of various LLMs based on 1 million tokens.}
\label{tab:model_prices}
\end{table}

\section{Case Study}
Figure~\ref{fig:case_study} illustrates a representative case on GPQA, highlighting how OI-MAS allocates agent role and model capacity based on the complexity of intermediate reasoning states. OI-MAS first activates \textsc{Decomposer} and \textsc{Generator} to structure the problem and produce an initial solution. In this case, \textsc{Decomposer} is routed to a lightweight backbone (Qwen2.5-3B) since it mainly performs straightforward categorization (e.g., identifying acids/bases and possible reaction types), while \textsc{Generator} is routed to a large model (Llama3.1-70B) because the initial quantitative setup (e.g., effective $H^+$/$OH^-$ amounts and the limiting condition) is the most error-sensitive step and largely determines the downstream trajectory. OI-MAS then invokes \textsc{Refiner} and \textsc{Verifier} jointly to revise the solution, routing \textsc{Refiner} to Qwen2.5-7B and \textsc{Verifier} to Llama3.1-8B. Medium-scale backbones are used for both roles, as this stage is more structured and check-oriented than the initial generation while still requiring reliable execution. Finally, OI-MAS applies \textsc{Ensembler} (Qwen2.5-7B) to consolidate the corrected reasoning paths and select the final answer, after which \textsc{EARLYSTOP} is triggered to terminate further interaction once the remaining uncertainty is resolved.

\end{document}